\documentclass[runningheads]{llncs}

\usepackage{graphicx}
\usepackage{comment}
\usepackage{amsmath,amssymb} 
\usepackage{color}

\usepackage{cuted}
\usepackage{capt-of,etoolbox}
\usepackage{hyperref}
\usepackage{epsfig}
\usepackage{graphicx}
\usepackage{bm}
\usepackage{multirow}

\newcommand{\tref}[1]{Tab.~\ref{#1}}

\newcommand{\fref}[1]{Fig.~\ref{#1}}
\newcommand{\sref}[1]{Sec.~\ref{#1}}

\newcommand{\tabincell}[2]{\begin{tabular}{@{}#1@{}}#2\end{tabular}}

\begin{document}
\pagestyle{headings}
\mainmatter
\def\ECCVSubNumber{2666}  

\title{Dense Hybrid Recurrent Multi-view Stereo Net with Dynamic Consistency Checking}

\titlerunning{$D^{2}$HC-RMVSNet}
%

\def\thefootnote{*}\footnotetext{Equal Contribution.}
\def\thefootnote{$\dagger$}\footnotetext{Corresponding Author}

\author{Jianfeng Yan\inst{1}$^*$ \and
Zizhuang Wei\inst{1}$^*$ \and
Hongwei Yi\inst{1}$^{*\dagger}$ \and
Mingyu Ding\inst{2} \and Runze Zhang\inst{3} \and 
Yisong Chen\inst{1} \and Guoping Wang\inst{1} \and
Yu-Wing Tai\inst{4}
}

\authorrunning{J. Yan, Z. Wei, H. Yi et al.}
%
\institute{
Peking University \\
\email{\{haibao637, weizizhuang, hongweiyi, chenyisong, wgp\}@pku.edu.cn} \and
HKU \hspace{2mm} \email{myding@cs.hku.hk} \and
Tencent \hspace{2mm}
\email{ryanrzzhang@tencent.com} \and
Kwai Inc. \hspace{2mm}
\email{yuwing@gmail.com}
}
\maketitle

\begin{abstract}
In this paper, we propose an efficient and effective dense hybrid recurrent multi-view stereo net with dynamic consistency checking, namely $D^{2}$HC-RMVSNet, for accurate dense point cloud reconstruction.
Our novel hybrid recurrent multi-view stereo net consists of two core modules: 1) a light DRENet (Dense Reception Expanded) module to extract dense feature maps of original size with multi-scale context information, 2) a HU-LSTM (Hybrid U-LSTM) to regularize 3D matching volume into predicted depth map, which efficiently aggregates different scale information by coupling LSTM and U-Net architecture.
To further improve the accuracy and completeness of reconstructed point clouds, we leverage a dynamic consistency checking strategy instead of prefixed parameters and strategies widely adopted in existing methods for dense point cloud reconstruction. In doing so, we dynamically aggregate geometric consistency matching error among all the views.
Our method ranks \textbf{$1^{st}$} on the complex outdoor \textsl{Tanks and Temples} benchmark over all the methods.
Extensive experiments on the in-door DTU dataset show our method exhibits competitive performance to the state-of-the-art method while dramatically reduces memory consumption, which costs only $19.4\%$ of R-MVSNet memory consumption. The codebase is available at \hyperlink{https://github.com/yhw-yhw/D2HC-RMVSNet}{https://github.com/yhw-yhw/D2HC-RMVSNet}.

\end{abstract}

\section{Introduction}
Dense point cloud reconstruction from multi-view stereo (MVS) information is a classic and important Computer Vision problem for decades, where stereo correspondences of more than two calibrated images are used to recover dense 3D representation~\cite{schonberger2016structure,lhuillier2005quasi,seitz2006comparison,strecha2008benchmarking,xu2019multi}.
While traditional MVS methods have achieved promising results, 
the recent advance in deep learning~\cite{flynn2016deepstereo,ji2017surfacenet,huang2018deepmvs,yao2018mvsnet,im2019dpsnet,yao2019recurrent,chen2019point,luo2019p} allows the exploration of implicit representations of multi-view stereo, hence resulting in superior completeness and accuracy in MVS benchmarks~\cite{aanaes2016DTU,knapitsch2017tanks} compared with traditional alternatives without learning.

However, those deep learning based MVS methods still have the following problems. 
First, due to the memory limitation, some methods like MVSNet~\cite{yao2018mvsnet} cannot deal with images with large resolutions. Then, RMVSNet~\cite{yao2019recurrent} are proposed to solve this problem, while the completeness and accuracy of reconstruction are compromised. 
Second, heavy backbones with downsampling module have to be used to extract features in~\cite{yao2018mvsnet,im2019dpsnet,yao2019recurrent,chen2019point,luo2019p}, which rely on large memory and lose information in the downsampling process.
At last, those deep learning based MVS methods have to fuse the depth maps obtained by different images. The fusion criteria are set in a heuristic pre-defined manner for all data-sets, which lead to low complete results.

To tackle the above problems, we propose a novel deep learning based MVS method called $D^{2}$HC-RMVSNet with a network architecture and a dynamic algorithm to fuse depth maps in the postprocessing. 
The network architecture consists of 1) a newly designed lightweight backbone to extract features for the dense depth map regression, 2) a hybrid module coupling LSTM and U-Net to regularize 3D matching volume into predicted depth maps with different level information into LSTM.
The dynamic algorithm to fuse depth maps attempts to aggregate the matching consistency among all the neighbor views to dynamically remain accurate and more reliable dense points in the final results.

Our main contributions are listed below:
\begin{itemize}
\item We propose a new lightweight DRENet to extract dense feature map for dense point cloud reconstruction.
\item We design a hybrid architecture DHU-LSTM which absorbs both the merits of LSTM and U-Net to reduce the memory cost while maintains the reconstruction accuracy.
\item We design a non-trivial dynamic consistency checking algorithm for filtering to remain more reliable and accurate depth values and obtain more complete dense point clouds.
\item Our method ranks \textbf{$1^{st}$} on \textsl{\textbf{Tanks and Temples}}~\cite{knapitsch2017tanks} among all methods and exhibits competitive performance to the state-of-the-art method on \textbf{DTU}~\cite{aanaes2016DTU} while dramatically reduces memory consumption.
\end{itemize}

\section{Related Work}

Deep neural network has made tremendous progress in many vision task~\cite{dosovitskiy2015flownet,kendall2017end,ding2020learning,yi2019mmface}, including several attempts on multi-view stereo. 
The deep learning based MVS methods~\cite{flynn2016deepstereo,ji2017surfacenet,huang2018deepmvs,yao2018mvsnet,im2019dpsnet,yao2019recurrent,chen2019point,luo2019p,yi2019pyramid} generally first use the backbones with some downsampling modules to extract features and the final layer of the backbones with the most downsampled feature maps are output to the following module. Hence, those methods cannot directly output depth maps with the same resolution as the input images and may lose some information in those higher resolutions, which may influence the accuracy of reconstructed results. 

Then, plane-sweep volumes are pre-wraped from images as the input to those networks~\cite{flynn2016deepstereo,ji2017surfacenet,huang2018deepmvs}. The plane-sweep volumes are memory-consuming and those methods cannot be trained end-to-end. To train the neural network in an end-end fasion, MVSNet~\cite{yao2018mvsnet} and DPSNet~\cite{im2019dpsnet} implicitly encodes multi-view camera geometries into the network to build the 3D cost volumes by introducing the differential homography warping. P-MVSNet~\cite{luo2019p} utilizes a patch-wise matching module to learn the isotropic matching confidence inside the cost volume.
PointMVSNet~\cite{chen2019point} proposes a two-stage coarse-to-fine method to generate high resolution depth maps, where a coarse depth map is first yielded by the lower-resolution version MVSNet~\cite{yao2018mvsnet} and depth errors are iteratively refined in the point cloud format. However, this method is time-consuming and complicated to employ in real applications since it consists of two different network architectures. 
In addition, the memory-consuming 3D-CNN modules adopted in those methods limit their application for scalable 3D reconstruction from high resolution images. To reduce memory consumption during the inference phase, R-MVSNet~\cite{yao2019recurrent} leverages the recurrent gated recurrent unit (GRU) instead of 3D-CNN, whereas compromises completeness and accuracy on 3D reconstruction.

All of the above methods have to fuse the depth maps from different reference images to obtain the final reconstructed dense point clouds by following the post-processing in the non-learning based MVS method COLMAP ~\cite{schonberger2016pixelwise}. In the post-processing, consistency is checked in a pre-defined manner, which is not robust for different scenes and may miss some good points viewed by few images.

To improve the deep learning based MVS methods based on above analysis, we propose a light DRENet specifically designed for the dense depth reconstruction, which outputs the same feature map size as input images with large receptive fields. Then a HU-LSTM (Hybrid U-LSTM) module is designed to reduce the memory consumption while maintains the 3D reconstruction accuracy. At last, we design a dynamic consistency checking algorithm for filtering to obtain more accurate and complete dense point clouds.

\section{Reconstruction Pipeline}
Given a set of multi-view images and corresponding calibrated camera parameters calculated from Structure-from-Motion~\cite{schoenberger2016sfm}, our goal is to estimate the depth map of each reference image and reconstruct dense 3D point cloud.
First, each input image is regarded as the reference image and fed to the effective Dense Hybrid Recurrent MVSNet (DH-RMVSNet) with several neighbor images to regress the corresponding dense depth map.
Then, we use a dynamic consistency checking algorithm to filter all the estimated depth maps of multi-view images to obtain more accurate and reliable depth values, by leveraging geometric consistency through all neighbor views.
After achieving dense filtered reliable depth maps, we directly re-project and fuse all pixels with reliable depth values into 3D space to generate corresponding dense 3D point clouds.

In the following sections, we first introduce our efficient DH-RMVSNet and novel dynamic consistency checking. Then we evaluate our method on DTU~\cite{aanaes2016DTU} and \textsl{Tanks and Temples}~\cite{knapitsch2017tanks} to prove the efficacy of our method.
To evaluate the practicality and generalization on the large-scale dataset with wide range for wide real application, we extend our method on aerial photos in the \textsl{BlendMVS}~\cite{yao2019blendedmvs} dataset to reconstruct a large-scale scene.

\section{Dense Hybrid Recurrent MVSNet}

\begin{figure}[t]
  \centering
  \includegraphics[width=\columnwidth]{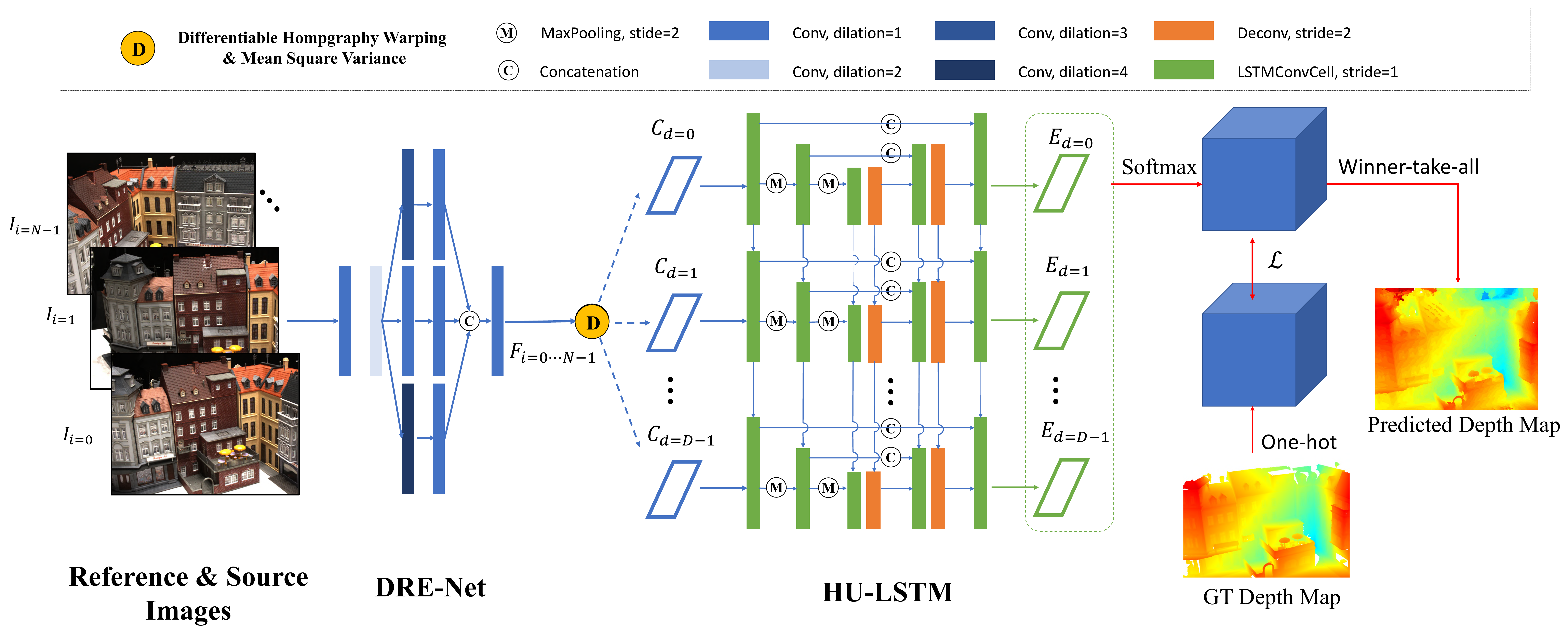}
  \caption{The network architecture of DH-RMVSNet. 2D feature maps extracted from multi-view images by DRENet go through differentiable homography warping and mean square variance to generate 3D cost volumes. Our HU-LSTM processes 3D cost volume sequentially in depth direction for further training or depth prediction.}
  \label{architecture}
\end{figure}

This section describes the details of our proposed network DH-RMVSNet as visualized in \fref{architecture}. 
We design a novel hybrid recurrent multi-view stereo network which absorbs both advantages of 3DCNN in MVSNet~\cite{yao2018mvsnet} and recurrent unit in R-MVSNet~\cite{yao2019recurrent}. 
Specifically, our DH-RMVSNet leverages well both the accuracy of 3DCNN processing 3D dimension data and the efficiency of recurrent unit by sequentially processing.
Therefore, our network can generate dense accurate depth maps and corresponding dense 3D reconstruction point clouds on the large-scale datasets.
We first introduce our lightweight efficient image feature extractor DRENet in~\sref{ednet}.
Then we present HU-LSTM sub-network to sequentially regularize feature matching volumes in the depth hypothesis direction into 3D probability volume in~\sref{DHR}. At last we introduce our training loss in~\sref{loss}.

\begin{table}[t]
    \centering
    \scriptsize
	\begin{tabular}{c|l|c|c}
	\hline
    Input & Layer Destription & Output & Output Size \\ \hline \hline
   \multicolumn{4}{c}{Input multi-view image size: $N\times H\times W\times 3$}\\ \hline \hline
    \multicolumn{4}{c}{DRENet} \\ \hline \hline
    $I_{i=0\cdots N-1}$ & ConvGR, filter=$3\times 3$, stride=$1$ & 2D0\_0 &  $H\times W\times 16$ \\
    2D0\_0 & ConvGR, filter=$3\times 3$, stride=$1$   & 2D0\_1 &  $H\times W\times 16$ \\
    2D0\_1 & ConvGR, filter=$3\times 3$, stride=$1$, dilation=2 &2D0\_2 &  $H\times W\times 32$ \\
    2D0\_2 & ConvGR, filter=$3\times 3$, stride=$1$  &2D0\_3&  $H\times W\times 32$ \\
    2D0\_2 & ConvGR, filter=$3\times 3$, stride=$1$, dilation=3  & 2D1\_1 &  $H\times W\times 32$ \\
    2D1\_1& ConvGR, filter=$3\times 3$, stride=$1$  & 2D1\_2& $H\times W\times 32$ \\
    2D0\_2& ConvGR, filter=$3\times 3$, stride=$1$, dilation=4 &2D2\_1 &  $H\times W\times 32$ \\
    2D2\_1& ConvGR, filter=$3\times 3$, stride=$1$  &2D2\_2 &  $H\times W\times 32$ \\ 
    $[$2D0\_3, 2D1\_2, 2D2\_2$]$& ConvGR, filter=$3\times 3$, stride=$1$  & $F_{i=0\cdots N-1}$ &  $H\times W\times 32$ \\
    \hline \hline
    \multicolumn{4}{c}{HU-LSTM} \\ \hline \hline
    $\mathcal{C}\left(i\right)$ & ConvLSTMCell, filter=$3\times 3$ & $\mathcal{C}_{0}\left(i\right)$ & $H\times W\times 32$ \\
   $ \mathcal{C}_{0}\left(i\right)$ & MaxPooling, stride=2 & $\mathcal{C}_{0}^{'}\left(i\right)$ & $\frac{1}{2}H\times  \frac{1}{2}W\times 32$ \\
    
   $ \mathcal{C}_{0}^{'}\left(i\right)\&\mathcal{C}_{2}\left(i-1\right)$ & ConvLSTMCell, filter=$3\times 3$ & $\mathcal{C}_{1}\left(i\right)$ & $H\times W\times 32$ \\
    
    $\mathcal{C}_{1}\left(i\right)$ & MaxPooling, stride=2 & $\mathcal{C}_{1}^{'}\left(i\right)$ & $\frac{1}{4}H\times  \frac{1}{4}W\times 32$ \\
    
    $\mathcal{C}_{1}^{'}\left(i\right)\&\mathcal{C}_{2}\left(i-1\right)$ & ConvLSTMCell, filter=$3\times 3$ & $\mathcal{C}_{2}\left(i\right)$ & $H\times W\times 32$ \\
    
    $\mathcal{C}_{2}\left(i\right)$ & DeConv, filter=$3\times 3$, stride=2 & $\mathcal{\hat{C}}_{2}\left(i\right)$ & $\frac{1}{2}H\times  \frac{1}{2}W\times 32$ \\
    
   $[$$\mathcal{C}_{1}\left(i\right),\mathcal{\hat{C}}_{2}\left(i\right)$$[$ \& $\mathcal{C}_{3}\left(i-1\right)$ & ConvLSTMCell, filter=$3\times 3$ & $\mathcal{C}_{3}\left(i\right)$ & $H\times W\times 32$ \\
    
   $ \mathcal{C}_{3}\left(i\right)$ & DeConv, filter=$3\times 3$, stride=2 & $\mathcal{\hat{C}}_{3}\left(i\right)$ & $\frac{1}{2}H\times  \frac{1}{2}W\times 32$ \\
    
    $[$$\mathcal{C}_{1}\left(i\right),\mathcal{\hat{C}}_{3}\left(i\right)$$]$\& $\mathcal{C}_{4}\left(i-1\right)$ & ConvLSTMCell, filter=$3\times 3$ & $\mathcal{C}_{4}\left(i\right)$ & $H\times W\times 32$ \\ \hline
    $\mathcal{C}_{4}\left(i\right)$ & Conv, filter=$3\times 3$, stride=$1$ & $\mathcal{C}_{H}\left(i\right)$ & 
    $H\times W\times 1$ \\ \hline \hline
    
	\end{tabular}
	\caption{The details of our DH-RMVSNet architecture which consists of DRENet and HU-LSTM. Conv and Deconv denote 2D convolution and 2D deconvolution respectively, GR is the abbreviation of group normalization and the ReLU. MaxPooling represents 2D max-pooling layer and ConvLSTMCell represent LSTM recurrent cell with 2D convolution. N, H, W, D are input multi-view number, image height, width and depth hypothesis number.}
	\label{dhrmvsnet_details}
\end{table}

\subsection{Image Feature Extractor}\label{ednet}
We design a Dense Receptive Expansion sub-network by concatenating feature maps from different dilated convolutional~\cite{yu2015multi} layers to aggregate multi-scale contextual information without losing resolution. We term it DRENet whose weights are shared by multi-view images $\textbf{I}_{i=0\cdots N-1}$.
Most of previous multi-view stereo network, such as \cite{yao2018mvsnet,yao2019recurrent,luo2019p}, usually use 2D convolutional layers with stride larger or equal than $2$ to enlarge the receptive field and reduce the resolution at the same time for satisfying memory limitation.
We introduce different dilated convolutional layers to generate multi-scale context information and preserve the resolution which leads to the possibility of dense depth map estimation. 
The details of DRENet are presented in \tref{dhrmvsnet_details}. 

Given $N$-view images, let $\textbf{I}_{i=0}$ and $\textbf{I}_{i=1\cdots N-1}$ denote the reference image and the neighbor source images respectively.
We first use two usual convolutional layers to sum up local-wise pixel information, then we utilize three dilated convolutional layers with different dilated ratio $2, 3, 4$ to extract multi-scale context information without scarifying resolution. 
Thus, after concatenation, DRENet can extract the dense feature map $F_{i}\in \mathbb{R}^{C\times H\times W}$ efficiently, where $C$ denotes the feature channel and $H, W$ represent the height and width of the input image.

Following common practices~\cite{yao2018mvsnet,im2019dpsnet,yao2019recurrent,chen2019point,luo2019p}, to build a 3D feature volume $\left\{\boldsymbol{V}_{i}\right\}_{i=0}^{N-1}$, we utilize the differentiable homography to warp the extracted feature map between different views. And we adopt the same mean square variance to aggregate them into one cost volume $\boldsymbol{C}$.

\subsection{Hybrid Recurrent Regularization}\label{DHR}
There exists two different ways to regularize the cost volume $\boldsymbol{C}$ into one probability map $\mathcal{P}$. One is to utilize the 3DCNN U-Net in MVSNet~\cite{yao2018mvsnet} which can well leverage local wise information and multi-scale context information, but it can not directly be used to regress the original dense depth map estimation due to limited GPU memory especially for large resolution images. The other is to use stacked convolutional GRU in R-MVSNet~\cite{yao2019recurrent} which is quiet efficient by sequentially processing the 3D volume through the depth direction but loss the aggregation of multi-scale context information.

Therefore, we absorb the merits in both two methods to propose a hybrid recurrent regularization network with more powerful recurrent convolutional cell than GRU, namely LSTMConvCell~\cite{xingjian2015convolutional}. We construct a hybrid U-LSTM which is a novel 2D U-net architecture where each layer is LSTMConvCell, which can be processed sequentially. We term this module HU-LSTM. Our HU-LSTM can well aggregate multi-scale context information and easily process dense original size cost volumes with high efficiency at the same time. It costs $19.4\%$ GPU memory of the previous recurrent method R-MVSNet~\cite{yao2019recurrent}. The detailed architecture of HU-LSTM is demonstrated in \tref{dhrmvsnet_details}.

Cost volume $\boldsymbol{C}$ can be viewed as $D$ number 2D cost matching map $\left\{\mathcal{C}\left(i\right)\right\}_{i=0}^{D-1}$ which are concatenated in the depth hypothesis direction. We denote the output of regularized cost matching map as $\left\{\mathcal{C}_{H}\left(i\right)\right\}_{i=0}^{D-1}$ at $i^{th}$ step during sequential processing. Therefore, $\mathcal{C}_{H}\left(i\right)$ relies on the both current input cost matching map $\mathcal{C}\left(i\right)$ and all previous states $\mathcal{C}_{H}\left(0,\cdots,i-1\right)$. Different from GRU in R-MVSNet~\cite{yao2019recurrent}, we introduce more powerful recurrent unit named ConvLSTMCell which has three gates map to control the information flow and can well aggregate different scale context information.

Let $\mathbb{I}\left(i\right)$, $\mathbb{F}\left(i\right)$ and $\mathbb{O}\left(i\right)$ denote the input gate map, forget gate map and output gate map respectively. In the following part, $\odot$, `[]' and `*' represent the element-wise multiplication, the concatenation and the matrix multiplication respectively in convolutional layer.

The input gate map is used to select valid information from current input $\mathcal{\hat{C}}\left(i\right)$ into the current state cell $\mathcal{C}\left(i\right)$:
\begin{equation}
    \mathbb{I}\left(i\right)=\sigma(\mathbb{W}_{\mathbb{I}}*[\mathcal{C}\left(i\right),\mathcal{C}_{H}\left(i-1\right)]+\mathbb{B}_{\mathbb{I}}), 
\end{equation}
\begin{equation}
    \mathcal{\hat{C}}\left(i\right)=\textsl{tanh}(\mathbb{W}_{\mathbb{C}}*[\mathcal{C}\left(i\right),\mathcal{C}_{H}\left(i-1\right)]+\mathbb{B}_{\mathbb{C}}),
\end{equation}
while the forget gate map $\mathbb{F}\left(i\right)$ decides to filter useless information from previous state cell $\mathcal{C}\left(i-1\right)$ and combines the input information from the input gate map $\mathbb{I}\left(i\right)$ to generate current new state cell $\mathcal{C}\left(i\right)$:
\begin{equation}
    \mathbb{F}\left(i\right)=\sigma(\mathbb{W}_{\mathbb{F}}*[\mathcal{C}\left(i\right),\mathcal{C}_{H}\left(i-1\right)]+\mathbb{B}_{\mathbb{F}}), 
\end{equation}
\begin{equation}
    \mathcal{C}\left(i\right)=\mathbb{F}_{i}\odot\mathcal{C}_{H}\left(i-1\right)+\mathbb{I}_{i}\odot \mathcal{\hat{C}}\left(i\right),
\end{equation}
Finally, the output gate map controls how much information from new current state cell $\mathcal{C}\left(i\right)$ will output, which is $\mathcal{C}_{H}\left(i\right)$ :
\begin{equation}
    \mathbb{O}\left(i\right)=\sigma(\mathbb{W}_{\mathbb{O}}*[\mathcal{C}\left(i\right),\mathcal{C}_{H}\left(i-1\right)]+\mathbb{B}_{\mathbb{O}}), 
\end{equation}
\begin{equation}
    \mathcal{C}_{H}\left(i\right)=\mathbb{O}\left(i\right)\odot\textsl{tanh}(\mathcal{C}\left(i\right)),
\end{equation}
where $\sigma$ and \textsl{tanh} represent \textsl{sigmoid} and \textsl{tanh} non-linear activation function respectively, $\mathbb{W}$ and $\mathbb{B}$ are learnable parameters in LSTM convolutional filter.

In our proposed HU-LSTM, by aggregating different scale context information to improve the robustness and accuracy of depth estimation, we adopt three LSTMConvCells to propragate different scale input feature maps with downsampling scale $0.5$ and two LSTMConvCell to aggregate multi-scale context information as denoted in \tref{dhrmvsnet_details}. Specifically, we input the $32$-channel input cost map $\mathcal{C}_{i}$ to the first LSTMConvCell, and the output of each LSTMConvCell will be fed into next LSTMConvCell. Then the regularized cost matching volume $\left\{\mathcal{C}_{H}\left(i\right)\right\}_{i=0}^{D-1}$ goes through by a \textsl{softmax} layer to generate corresponding the probability volume $\mathcal{P}$ for further caculating training loss.

\subsection{Training Loss}\label{loss}
Following MVSNet~\cite{yao2018mvsnet}, we
treat the depth regression task as multiple classification task and use the same cross entropy loss function $\mathcal{L}$ between the probability volumes $\mathcal{P}$ and ground truth depth map $\mathcal{G}$:
\begin{equation}
\mathcal{L}=\sum_{\boldsymbol{x} \in \boldsymbol{x}_{valid}}\sum_{i=0}^{D-1}-G(i,x)*log(P(i,x)), 
\end{equation}
where $\boldsymbol{x}_{valid}$ is the set of valid pixels in the ground truth, $G(i,x)$ represents the one-hot vector generated by the depth value of the ground truth $\mathcal{G}$ at pixel $x$ and $P(i,x)$ is the corresponding depth estimated probability.
During test phase, we do not need to save the whole probability map. To further improve the efficiency, the depth map is processed sequentially and the winner-take-all selection is used to generate the estimated depth map from regularized cost matching volume.

\section{Dynamic Consistency Checking}
The above DH-RMVSNet generates dense pixel-wise depth map for each input multi-view images. 
Before fusing all the estimated multi-view depth maps, it is necessary to filter out mismatched errors and store correct and reliable depths.
All previous methods~\cite{yao2018mvsnet,yao2019recurrent,chen2019point,luo2019p} just follow \cite{schoenberger2016mvs} to apply the geometric constraint to measure the depth estimation consistency among multiple views.
However, those methods only use prefixed constant parameters. Specifically, the reliable depth value should satisfy both conditions: the pixel reprojection error less than $\tau_{1}$ and and the depth reprojection error less than $\tau_{2}$ in at least three views, where $\tau_{1}=1$ and $\tau_{2}=0.01$ are pre-defined. 
These parameters are defined intuitively and not robust for different scenes, though they have large influence on the quality of reconstruct point cloud. 
For example, those depth values with much high reliable consistency in two views are filtered and 
a fixed number valid views also lose information in the views with slightly worse errors. 
Beside, using the fixed $\tau_{1}$ and $\tau_{2}$ may not filter enough mismatched pixels in different scenes.

In general, a estimated depth value is accurate and reliable when it has a very low reprojection error in few views, or a lower error in majority views.
Therefore, we propose a novel dynamic consistency checking algorithm to select valid depth values, which is related to both the reprojection error and view numbers. 
By considering dynamic geometric matching cost as consistency among all neighbor views, it leads to more robust and complete dense 3D point clouds. 
We denote the estimated depth value $D_i(\boldsymbol{p})$ of a pixel $\boldsymbol{p}$ on reference image $\textbf{I}_i$ through our DH-RMVSNET.
The camera parameter is represented by $\boldsymbol{P}_i=[\boldsymbol{M}_i|\boldsymbol{t}_i]$ in~\cite{hartley2003multiple}.
First we back-project the pixel $p$ into 3D space to generate the corresponding 3D point $\boldsymbol{X}$ by:   
\begin{equation}
    \boldsymbol{X} = \boldsymbol{M}_i^{ - 1}({\boldsymbol{D}_i}(\boldsymbol{p}) \cdot \boldsymbol{p} - {\boldsymbol{t}_i}),\end{equation}
Then we project the 3D point $\boldsymbol{X}$ to generate the projected pixel $\boldsymbol{q}$ on the neighbor view $\textbf{I}_j$:
\begin{equation}
  \boldsymbol{q} = \frac{1}{d}{\boldsymbol{P}_j} \cdot \boldsymbol{X},
\end{equation}
where $\boldsymbol{P}_j$ is the camera parameter of neighbor view $\textbf{I}_j$ and $d$ is the depth from projection.
In turn, we back-project the projected pixel $\boldsymbol{q}$ with estimated depth $D_j(\boldsymbol{q})$ on the neighbor view into 3D space and reproject back to the reference image denoted as $\boldsymbol{p}'$:
\begin{equation}
  \boldsymbol{p}' = \frac{1}{{d'}}{\boldsymbol{P}_j} \cdot (\boldsymbol{M}_j^{ - 1}({D_j}(\boldsymbol{q}) \cdot \boldsymbol{q} - {\boldsymbol{t}_j})),
\end{equation}
where $d'$ is the depth value of the reprojected pixel $\boldsymbol{p}'$ on the reference image.
Based on the above mentioned operation, the reprojection errors are calculated by:
\begin{equation}
   \begin{array}{l}
{\xi _p} = ||\boldsymbol{p} - \boldsymbol{p}'|{|_2},\\
{\xi _d} = ||{D_i}(\boldsymbol{p}) - {d'}|{|_1}/{D_i}(\boldsymbol{p}).
  \end{array}
\end{equation}
In order to quantify the depth matching consistency between two different views, we propose the dynamic matching consistency by considering dynamic matching consistency among all views. The dynamic matching consistency in different views is defined as:
\begin{equation}
   {c_{ij}(\boldsymbol{p})} = {e^{ - ({\xi _p} + \lambda  \cdot {\xi _d})}},
\end{equation}
where $\lambda$ is used to leverage the reprojection error in two different metrics. By aggregating the matching consistency from all the neighbor views to obtain the global dynamic multi-view geometric consistency $C_{geo}(\boldsymbol{p})$ as:
\begin{equation}
   {C_{geo}}(\boldsymbol{p}) = \sum\limits_{j = 1}^N {{c_{ij}}}.
\end{equation}

We calculate the dynamic geometric consistency for every pixel and filter out the outliers with ${C_{geo}}(\boldsymbol{p})<\tau$.
Benefiting from our proposed dynamic consistency checking algorithm, 
the filtered depth map is able to store more accurate and complete depth values compared with the previous intuitive fixed-threshold method. It improves the robustness, completeness and accuracy of 3D reconstructed point clouds.

\section{Experiments}
\subsection{Implementation Details}
\paragraph{Training}
We train DH-RMVSNet on the DTU dataset~\cite{aanaes2016DTU}, which contains 124 different indoor scenes
which is split to three parts, namely \textsl{training}, \textsl{validation} and \textsl{evaluation}.
Following common practices~\cite{huang2018deepmvs,ji2017surfacenet,yao2018mvsnet,yao2019recurrent,chen2019point,yi2019pyramid,ChenPMVSNet2019ICCV}, we train our network on the training dataset and evaluate on the evaluation dataset. 
While the dataset only provides ground truth point clouds generated by scanners, to generate the ground truth depth map, we use the same rendering method as MVSNet~\cite{yao2018mvsnet}.
Different from \cite{yao2018mvsnet,yao2019recurrent} which generate the depth map with $\frac{1}{4}$ size of original input image, our method generates the depth map with the same size as input image. Since MVSNet~\cite{yao2018mvsnet} only provides $\frac{1}{4}$ size depth map, thus we resize the training input image to $W\times H = 160\times 128$ as same as the corresponding groundtruth depth map. 
We set the number of input images $N=3$ and the depth hypotheses are sampled from $425mm$ to $745mm$ with depth plane number $D=128$ in MVSNet~\cite{yao2018mvsnet}. 
We implement our network on \textbf{PyTorch}~\cite{paszke2017automatic} and train the network end-to-end for $6$ epochs using \textsl{Adam}~\cite{kingma2014adam} with an initial learning rate $0.001$ which is decayed by 0.9 every epoch. 
Batch size is set to $6$ on $2$ NVIDIA TITAN RTX graphics cards.
\paragraph{Testing.}\label{testing}
For testing, we use the $N=7$ views as input, and set $D=256$ for depth plane hypothesis in an inverse depth manner in~\cite{yao2019recurrent}. To evaluate \textsl{Tanks and Temples} dataset, the camera parameters are computed by OpenMVG~\cite{moulon2014openmvg} following MVSNet~\cite{yao2018mvsnet} and the input image resolution is set to $1920\times1056$. We test the BlendedMVS~\cite{yao2019blendedmvs} dataset using original images of $768\times 576$ resolution.

\paragraph{Filter \& Fusion}
After we generate estimated depth maps from DH-RMVSNet, we filter and fuse them to generate corresponding 3D dense point cloud.
First, the depths with probability lower than $\phi=0.4$ will be discarded.
Then, we use our proposed dynamic global geometric consistency checking algorithm as further multi-view depth map filter with $\lambda=200$ and $\tau=1.8$. At last, we fuse all reliable depths into 3D space to generate 3D point cloud.

\begin{table}[t]
    \centering
    \small
    \begin{tabular}{lccc}
    \hline
    \multirow{2}{*}{Method} & \multicolumn{3}{c}{Mean Distance (mm)} \\ 
    & Acc. & Comp. & \textsl{overall} \\ 
    \hline
    Tola \cite{tola2012efficient} & 0.342 & 1.190 & 0.766 \\
    Gipuma \cite{galliani2015massively} & \textbf{0.283} & 0.873 & 0.578 \\
    Colmap \cite{schonberger2016pixelwise} & 0.400 & 0.664 & 0.532 \\ 
    SurfaceNet \cite{ji2017surfacenet} & 0.450 & 1.040 & 0.745 \\ 
    MVSNet \cite{yao2018mvsnet} & 0.396 & 0.527 & 0.462 \\ 
    R-MVSNet \cite{yao2019recurrent} & 0.385 & 0.459 & 0.422 \\ 
    P-MVSNet \cite{luo2019p} & 0.406 & 0.434 & 0.420 \\
    PointMVSNet \cite{chen2019point} & 0.361 & 0.421 & 0.391 \\ 
    PointMVSNet-HiRes \cite{chen2019point} & 0.342 & 0.411 & \textbf{0.376} \\ 
     \hline
    \textbf{${\boldsymbol{D^{2}}}$HC-RMVSNet}  &0.395 & \textbf{0.378}   & 0.386  \\
    \end{tabular}
    \caption{Quantitative results on the DTU evaluation dataset~\cite{aanaes2016DTU} (lower is better). Our method $D^{2}$HC-RMVSNet exhibits a competitive reconstruction performance compared with state-of-the-art methods in terms of completeness and overall quality.}
    \label{eval_dtu}
\end{table}

\subsection{Datasets and Results}
We first demonstrate the state-of-the-art performance of our proposed $D^{2}$HC-RMVSNet on the DTU~\cite{aanaes2016DTU} and \textsl{Tanks and Temples}~\cite{knapitsch2017tanks}, which outperforms its original methods, namely MVSNet~\cite{yao2018mvsnet} and R-MVSNet~\cite{yao2019recurrent} with a significant margin. 
Specifically, our method ranks \textbf{$1^{st}$} in the complex large-scale outdoor \textsl{Tanks and Temples} benchmark over all existing methods. 
To investigate the practicality and scalability of our method, we extend our method on the aerial photos in \textsl{BlendedMVS}~\cite{yao2019blendedmvs} to reconstruct a larger scale scenes.

\begin{figure}[t]
 \centering
 \includegraphics[width=\columnwidth]{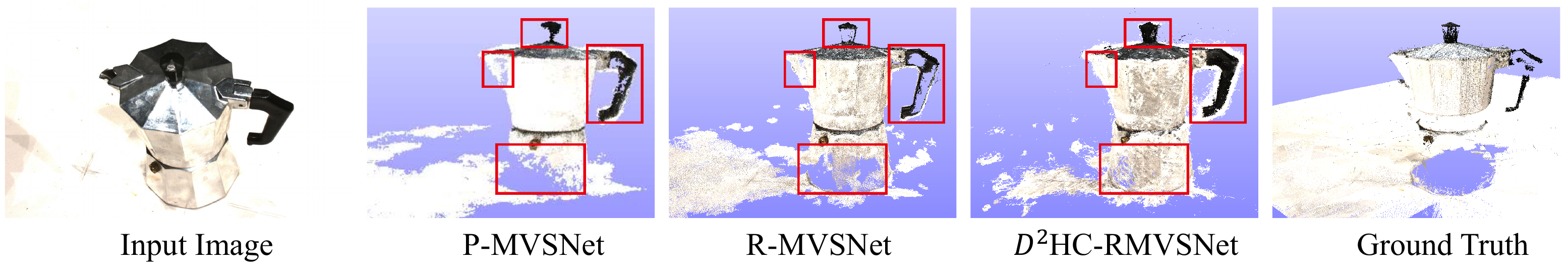}
 \caption{Comparison on the reconstructed point clouds for the \textsl{Scan} $77$ from the DTU~\cite{aanaes2016DTU} dataset with other methods~\cite{luo2019p,yao2019recurrent}. Our method generates more complete and denser point cloud than other methods.}
 \label{dtu_cmp}
\end{figure}

\paragraph{DTU Dataset.}
We evaluate our proposed method on the DTU~\cite{aanaes2016DTU} \textsl{evaluation} dataset.
We set $D=256$ within the depth range [425mm, 905mm] for all scans and use the common evaluation metric in other methods~\cite{yao2018mvsnet,yao2019recurrent}.
Quantitative results are shown in ~\tref{eval_dtu}. 
While Gipuma~\cite{galliani2015massively} performs the best regarding to accuracy, our method achieves the best completeness and the competitive \textsl{overall} quality of reconstruction results. 
Our proposed $D^{2}$HC-RMVSNet can both improve the accuracy and the completeness significantly compared with its original methods MVSNet~\cite{yao2018mvsnet} and R-MVSNet~\cite{yao2019recurrent}.
We also compare the results on the reconstructed point clouds with \cite{luo2019p,yao2019recurrent}. As shown in \fref{dtu_cmp}, our method generates more complete and accurate point cloud than other methods. It proves the efficacy of our novel DH-RMVSNet and dynamic consistency checking algorithm.

\begin{table*}[t]
    \centering
    \scriptsize
    \begin{tabular}{l|cccccccccc}
    \hline
    Method & Rank & Mean & Family & Francis & Horse & L.H. & M60 & Panther & P.G. & Train \\ \hline
    COLMAP~\cite{schoenberger2016sfm,schoenberger2016mvs}&	55.62&	42.14	&	50.41&	22.25&	25.63&	56.43&	44.83&	46.97&	48.53&	42.04\\
    Pix4D~\cite{pix4d}&	53.38&	43.24	&	64.45	&31.91&	26.43&	54.41	&50.58&	35.37&	47.78	&34.96\\
    MVSNet~\cite{yao2018mvsnet} &	52.75&	43.48&	55.99&	28.55&	25.07&	50.79&	53.96&	50.86&	47.90&	34.69\\
   Point-MVSNet~\cite{chen2019point}&	40.25&	48.27	&	61.79&	41.15&	34.20	&50.79&	51.97&	50.85&	52.38	&43.06\\
  Dense R-MVSNet~\cite{yao2019recurrent}&	37.50&	50.55&	73.01&	54.46&	43.42&	43.88	&46.80&	46.69&	50.87&	45.25\\ 
   OpenMVS~\cite{openmvs} &	17.88&	55.11	&	71.69&	51.12&	42.76&	58.98&	54.72&	56.17&	59.77&	45.69\\
  P-MVSNet~\cite{luo2019p}&	17.00&	55.62&	70.04&	44.64&	40.22&	\textbf{65.20}&	55.08&	55.17&	60.37&	54.29 \\ 
    CasMVSNet~\cite{gu2019cascade}&	14.00	&56.84	&	76.37&	58.45&	46.26&	55.81&	56.11&	54.06&	58.18&	49.51\\
    ACMM~\cite{xu2019multi} &	12.62	&57.27	&	69.24	&51.45	&46.97	&63.20&	55.07&	57.64&	\textbf{60.08}&	\textbf{54.48} \\
    Altizure-HKUST-2019~\cite{altizure}&	9.12&	59.03	&	\textbf{77.19} & \textbf{61.52} & 42.09 & 63.50& 59.36&	58.20&	57.05&	53.30\\ \hline
    \textbf{DH-RMVSNet} & 10.62 &	57.55& 73.62&	53.17&	46.24&	58.68&	59.38&	58.31&	58.26&	52.77 \\
    \textbf{${\boldsymbol{D^{2}}}$HC-RMVSNet} & \textbf{5.62}&	\textbf{59.20}	&	74.69&	56.04&	\textbf{49.42}&	60.08&	\textbf{59.81}&	\textbf{59.61}&	60.04&	53.92\\
    \hline
    \end{tabular}
    \caption{Quantitative results on the \textsl{Tanks and Temples} benchmark~\cite{knapitsch2017tanks}. The evaluation metric is \textsl{f-score} which higher is better. (L.H. and P.G. are the abbreviations of \textsl{Lighthouse} and \textsl{Playground} dataset respectively. )}
    \label{tat_eval}
\end{table*}

\begin{figure}[t]
    \centering
    \includegraphics[width=0.98\textwidth]{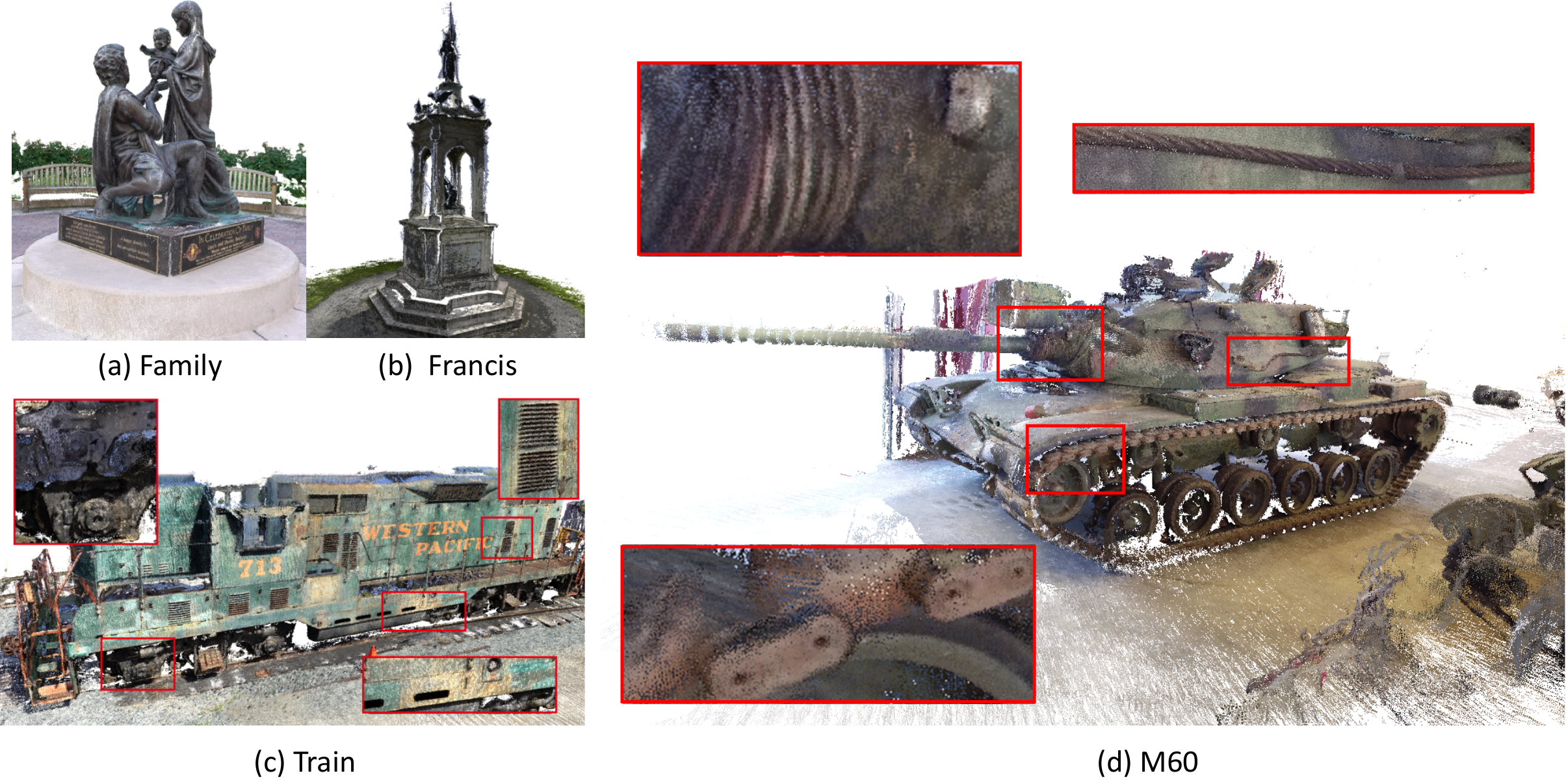}
    \caption{Point cloud results on the \textsl{Tanks and Temples} \cite{knapitsch2017tanks} benchmark, our method generates accurate, delicate and complete reconstructed point clouds which show the strong generalization of our method on complex outdoor scenes.}
    \label{TNT_results}
\end{figure}

\paragraph{Tanks and Temples Benchmark.}
\textsl{Tanks and Temples Benchmark}~\cite{knapitsch2017tanks} is a large-scale outdoor dataset which consists of more complex environment, and it is quite typical for real captured situation, compared with DTU dataset which is taken under well-controlled environment with fixed camera trajectory.

\begin{figure}[htbp]
    \centering 
    \includegraphics[width=0.49\columnwidth]{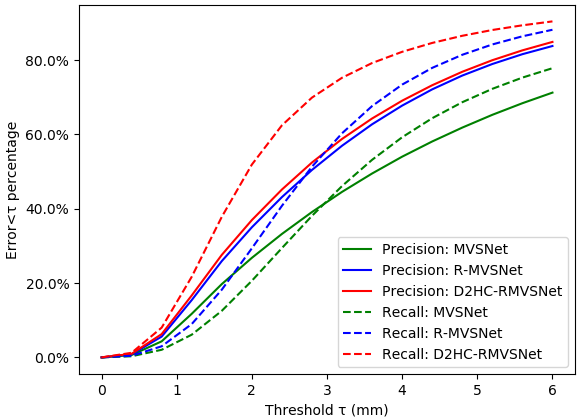}
    \caption{Comparison on the Precision / Recall (in\%) at different thresholds (within 6mm) on the \textsl{Family} provided by~\cite{tnt} with MVSNet~\cite{yao2018mvsnet} and R-MVSNet~\cite{yao2019recurrent}.}
    \label{tnt_pr}
\end{figure}

We evaluate our method \textbf{without any fine-tuning} on the \textsl{Tanks and Temples} as denoted in \tref{tat_eval}.
Our proposed $D^{2}$HC-RMVSNet ranks $1^{st}$ over all existing methods. Specifically, our method outperforms all deep-learning based multi-view stereo methods with a big margin. It shows the stronger generalization compared with Point-MVSNet~\cite{chen2019point} while Point-MVSNet~\cite{chen2019point} is the state-of-the-art method on the DTU~\cite{aanaes2016DTU}.
The mean \textsl{f-score} increases significantly from $50.55$ to $59.20$ (larger is better, date: Mar. 5, 2020) compared with Dense R-MVSNet~\cite{yao2019recurrent}, which demonstrates the efficacy and robustness of $D^{2}$HC-RMVSNet on the variant scenes.
The reconstructed point clouds are shown in \fref{TNT_results}, it shows that our method generates accurate, delicate and complete point cloud. And we compare the Precision / Recall of the model \textsl{Family} with its original methods~\cite{yao2018mvsnet,yao2019recurrent} at different error threshold in \fref{tnt_pr}. It demonstrates our method achieves a significant improvement on the precision while maintains better recall than R-MVSNet~\cite{yao2019recurrent}, which leads to the best performance on the \textsl{Tanks and Temples}.

\begin{figure}[t]
    \centering
    \includegraphics[width=\textwidth]{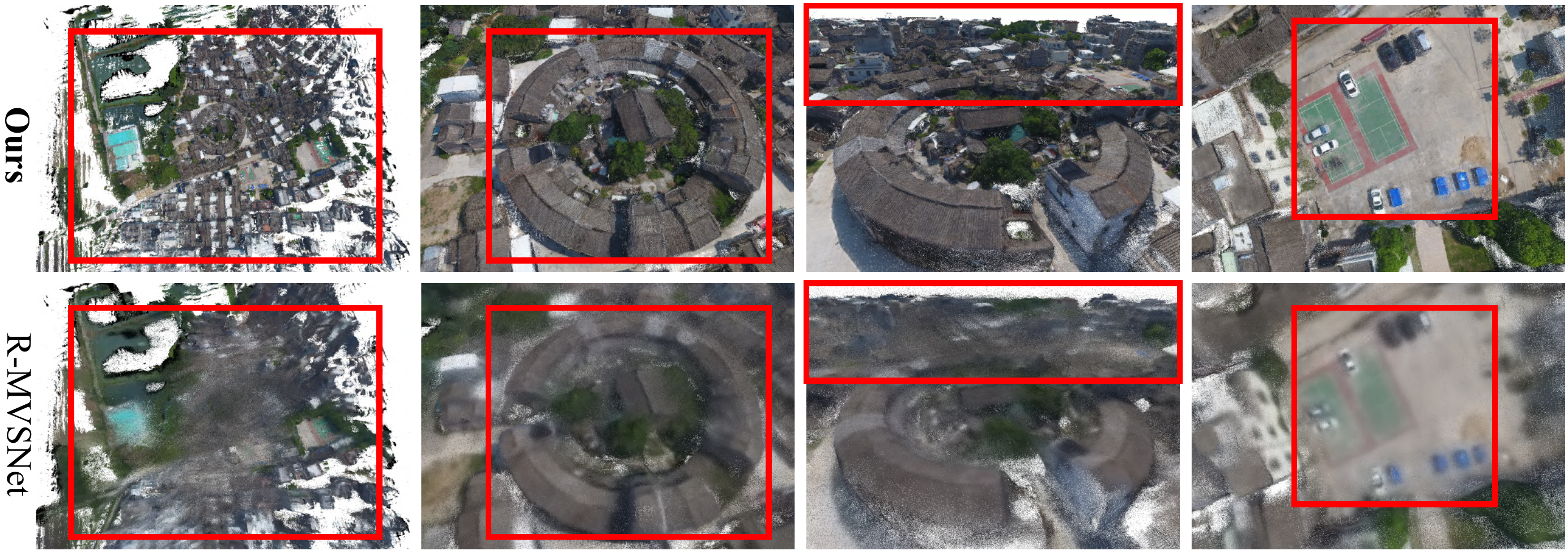}
    \caption{Comparison of the reconstruction point cloud results on the \textsl{validation} set of \textsl{BlendedMVS}~\cite{yao2019blendedmvs}. Our method can both well reconstruct a large-scale scene and small cars in it, while R-MVSNet~\cite{yao2019recurrent} failed on it. \textsl{(Both methods are without fine-tuning.)}}
    \label{blended_results}
\end{figure}

\paragraph{BlendedMVS.}
BlendedMVS is a new large-scale MVS dataset which is synthesized from 3D reconstructed models from Altizure~\cite{altizure}. The dataset contains over $113$ different scenes with a variety of different camera trajectories. And each scenes consists of 20 to 1000 input images including architectures, sculptures and small objects. To further evaluate the practicality and scalability of our propose $D^{2}$HC-RMVSNet, we directly test our method and R-MVSNet~\cite{yao2019recurrent} on the provided \textsl{validation} dataset.
For fair comparison, both methods are trained on the DTU~\cite{aanaes2016DTU} ~\textbf{without any finetuning} and we upsample the $\frac{1}{4}$ depth map from R-MVSNet to the same size of the depth map from our method, which is the original size of input images. 
As shown in~\fref{blended_results}, our method can well reconstruct the whole large scale scene and small cars in it, while R-MVSNet~\cite{yao2019recurrent} fails on it. 
Our method can estimate the dense delicate accurate depth map with original size of input image in an inverse depth setting as in~\cite{yao2019recurrent}, because of our novel DRENet and HU-LSTM, which has more accuracy and stronger scalability of 3D point cloud reconstruction by aggregating multi-scale context information on the large-scale dataset. Due to our dynamical consistency checking algorithm, we can directly use our algorithm to remain dense reliable point cloud without any specific adjustment.

\subsection{Ablation Study}
In this section, we provide ablation experiments to analyze the strengths of the key components of our architecture. we perform the following studies on DTU \textsl{validation} dataset with same setting in~\sref{testing}.

\paragraph{Variant components of network architecture}
To quantitatively analyze how different network architecture in DH-RMVSNet affect the depth map reconstruction, 
we evaluate the average mean absolute error between estimated depth maps and the ground truth on the \textsl{validation} DTU dataset during training. The comparison results are illustrated in~\fref{depth_error}.

We replace our ``DRENet'' and ``HU-LSTM'' with ``2DCNNFeatNet'' and  ``3D GRU'' in R-MVSNet~\cite{yao2019recurrent} respectively to analysis the influence of our proposed feature encoder and cost volume regularization module.
Comparied with ``2DCNNFeatNet + HU-LSTM'' in~\fref{depth_error}, our proposed ``DRENet'' can improve the accuracy slightly but with less inference time and memory consumption due to the light architecture. 
``HU-LSTM'' achieves significant improvement with a big margin compared with ``DRENet + 3D GRU'', which absorbs both the merits of efficacy in~\cite{yao2018mvsnet} and efficiency in~\cite{yao2019recurrent} by aggregating multi-scale context information in sequential process. 
To further evaluate the difference in two different powerful recurrent gate units, LSTM and GRU, we replace the LSTM cell in our ``HU-LSTM'' with ``GRU'', denoted as ``HU-GRU''. The comparison between ``HU-LSTM'' and ``HU-GRU'' shows ``LSTM'' is more accurate and robust than ``GRU'' because of more gate map to control the information flow and have a better performance on the learning matching patterns.


\makeatletter
\newcommand\figcaption{\def\@captype{figure}\caption}
\newcommand\tabcaption{\def\@captype{table}\caption}
\makeatother
\begin{figure}[t]
    \begin{minipage}[t]{.5\columnwidth}
    \centering
    \figcaption{Validation results of the mean average depth error with different network architectures during training.}
    \includegraphics[width=0.5\columnwidth]{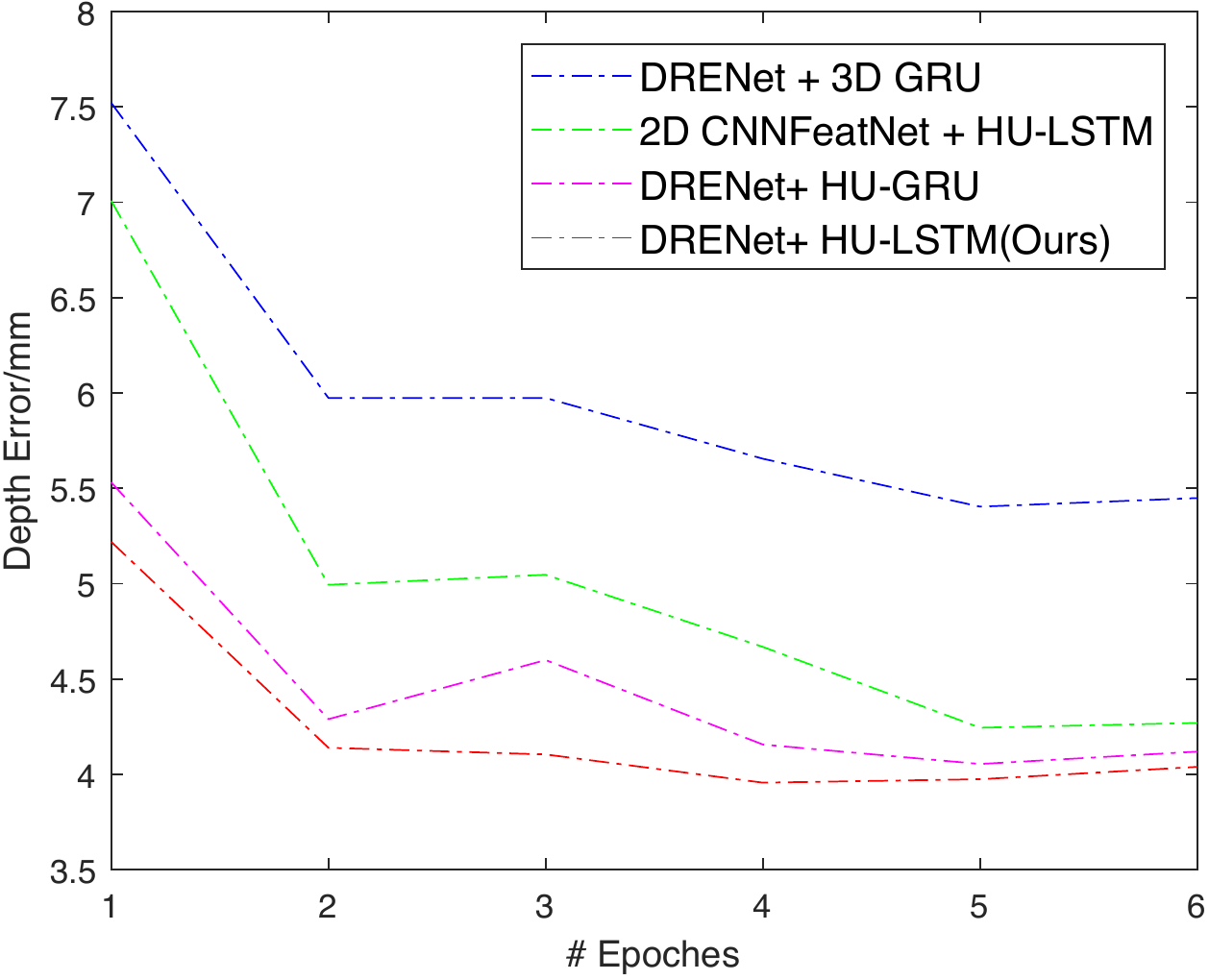}
    \label{depth_error}
    \end{minipage}
    \begin{minipage}[t]{.48\columnwidth}
    \centering
    \scriptsize
    \tabcaption{Comparison of the running time and memory consumption between our proposed $D^{2}$HC-RMVSNet and R-MVSNet~\cite{yao2019recurrent} on the DTU~\cite{aanaes2016DTU}.}
    \begin{tabular}{l|cccc}
    \hline
    \multirow{2}{*}{Method} & \multirow{2}{*}{Input Size} & \multirow{2}{*}{Output Size} & \multirow{2}{*}{\tabincell{c}{Mem.\\ $\left(GB\right)$}} & \multirow{2}{*}{\tabincell{c}{Time\\ $\left(s\right)$}} \\ 
      & & & & \\ \hline
    R-MVSNet & 1600x1196 & 400x296 & 6.7 & \textbf{2.1}\\ \hline
    \textbf{Ours} & 400x296 & 400x296 & \textbf{1.3} & 2.6   \\
    \textbf{Ours} & 800x600 & 800x592 &  2.4 & 8.0 \\
    \textbf{Ours} & 1600x1200 & 1600x1196 & 6.6 & 29.15  \\
    \hline
    \end{tabular}
    \label{memory}
    \end{minipage}
\end{figure}


\paragraph{Benefit from Dynamic Consistency Checking}
To further study the influence and generalization of our dynamic consistency checking, we evaluate our DH-RMVSNet with common filtering algorithm as in previous methods~\cite{yao2018mvsnet,yao2019recurrent,chen2019point} as shown in \tref{tat_eval}. Our proposed dynamic consistency checking algorithm significantly boosts the reconstruction results in all scenes on the \textsl{Tanks and Temples} benchmark, which shows the strong generalization and dynamic adaptation on the different scenes. It improves the \textsl{f-score} from $57.55$ by DH-RMVSNet to $59.20$, which leads to more accurate and complete reconstruction point clouds.



\section{Discussion}

\paragraph{Running Time \& Memory Utility}
For fair comparison on the running time and memory utility with R-MVSNet~\cite{yao2019recurrent}, we test our method with same depth sample number $D=256$ on the GTX 1080Ti GPU. As shown in~\tref{memory}, our method inputs multi-images of only $400\times296$ resolution to generate the depth map of the same size as R-MVSNet~\cite{yao2019recurrent}, with only $19.4\%$ memory consumption of R-MVSNet. Moreover, our method runs with $2.6\/s$ per view, with a little extra inference time than R-MVSNet, which needs an extra $6.2s$ refinement to enhance the performance. Our $D^{2}$HC-RMVSNet achieves significant improvement over R-MVSNet~\cite{yao2019recurrent} both on the DTU and the \textsl{Tanks and Temples} benchmark, while our novel dynamic consistency checking takes negligible running time.
Our method can generate dense depth maps with the same size of the input image with efficient memory consumption. It takes only $6.6GB$ to process multi-view images with $1600\times 1200$ resolution, which leads to a wide practicality for dense point cloud reconstruction.

\paragraph{Scalability and Generalization}
Due to our light DRENet and HU-LSTM, our method shows more powerful general scalability than R-MVSNet~\cite{yao2019recurrent} on the dense reconstruction with wide range. 
Our method can easily extend to aerial photos for the reconstruction of the big scene architectures in~\fref{blended_results} and generate denser, more accurate and complete 3D point cloud reconstruction due to the original size depth map estimation from our $D^{2}$HC-RMVSNet.


\section{Conclusions}
We have presented a novel dense hybrid recurrent multi-view stereo network with dynamic consistency checking, denoted as $D^{2}$HC-RMVSNet, for dense accurate point cloud reconstruction. 
Our DH-RMVSNet well absorbs both the merits of the accuracy of 3DCNN and the efficiency of Recurrent unit, to design a new lightweight feature extractor DRENet and hybrid recurrent regularization module HU-LSTM. 
To further improve the robustness and completeness of 3D point cloud reconstruction, we propose a no-trivial dynamic consistency checking algorithm to dynamically aggregate geometric matching error among all views rather than use prefixed strategy and parameters. 
Experimental results show that our method ranks $1^{st}$ on the complex outdoor \textsl{Tanks and Temples} and exhibits the competitive results on the $DTU$ dataset,
while dramatically reduces memory consumption, which costs only $19.4\%$ of R-MVSNet memory consumption. 

\noindent\textbf{Acknowledgements} 
This project was supported by the National Key R\&D Program of China (No.2017YFB1002705, No.2017YFB1002601) and NSFC of China (No.61632003, No.61661146002, No.61872398).
%
%
\bibliographystyle{splncs04}
\bibliography{egbib}
\end{document}